\documentclass{article}
\usepackage{spconf,amsmath,graphicx,hyperref}
\usepackage{tikz,pgfplots,tkz-berge}
\pgfplotsset{compat=1.18}

\title{Sparse Polyak with optimal thresholding operators for high-dimensional M-estimation}
\name{Tianqi Qiao, Marie Maros}
\address{Wm Michael Barnes '64 Industrial and Systems Engineering \\Texas A\&M university\\ College Station, TX, USA}
\usepackage{amsmath, amssymb, amsthm}
\usepackage{xcolor}

\usepackage{empheq}

\usepackage{algorithm}   
\usepackage{algpseudocode}

\usepackage{booktabs}

\newtheorem{assumption}{Assumption}

\newtheorem{theorem}{Theorem}

\newtheorem{corollary}{Corollary}

\newcommand{\hs}{\Phi^{\mathrm{HT}}_s}
\newcommand{\hts}{\Phi^{\mathrm{HT}}_{2s}}
\newcommand{\hxs}{\Phi^{\mathrm{HT}}_{s^*}}

\begin{document}
%
\maketitle
\begin{abstract}
We propose and analyze a variant of Sparse Polyak \cite{spolyak} for high dimensional M-estimation problems. Sparse Polyak proposes a novel adaptive step-size rule tailored to suitably estimate the problem's curvature in the high-dimensional setting, guaranteeing that the algorithm's performance does not deteriorate when the ambient dimension increases. However, convergence guarantees can only be obtained by sacrificing solution sparsity and statistical accuracy. In this work, we introduce a variant of Sparse Polyak that retains its desirable scaling properties with respect to the ambient dimension while obtaining sparser and more accurate solutions.
\end{abstract}
\vspace{-3mm}
\section{Introduction}
\label{sec:intro}
\vspace{-3mm}
We consider M-estimation problems of the form 
\vspace{-3mm}
\begin{align}\label{eq:l0}
\min_{\theta: \|\theta\|_{0} \leq s} \quad f(\theta) = \sum_{i=1}^n \ell (\theta, z_i) ,\\[-8mm]
\nonumber
\end{align}
where $z_i = (x_i,y_i),$ $i=1,\hdots,n,$  $x_i \in \mathbb{R}^d$ denote the feature vectors and $y_i \in \mathbb{R}$ denote the responses. We consider the high-dimensional regime, in which the ambient dimension $d$ is allowed to be much larger than the sample size $n$ and even, $\frac{d}{n} \to \infty.$  High-dimensional M-estimation problems of the form \eqref{eq:l0}, e.g., $\ell_0$ constrained linear or logistic regression, have emerged as a central topic in modern statistics and machine learning~\cite{nega,khanna2018,zhou2018}. In this context, the goal is to establish (near) optimal final precision, while guaranteeing fast convergence of computational methods. \cite{fast} established such guarantees for projected gradient descent when solving the LASSO. \cite{loh} and \cite{nega} extended the results in \cite{fast}  to a broader class of regularizers (e.g., MCP and SCAD) and generalized linear models. \cite{xiaolin,wang} used a homotopy continuation approach to guarantee the sparsity of solution. Jain et al. \cite{jain} investigated iterative hard thresholding (IHT), which emerges from applying projected gradient descent to \eqref{eq:l0}, and proved linear convergence guarantees. IHT takes a gradient step and then applies the hard thresholding operator:  
\begin{align}\label{eq:classic_iht}
    \theta_{t+1} = \hs (\theta_t - \gamma \nabla f(\theta_t))
\end{align}
making it desirable due to its simplicity. The hard thresholding operator $\hs$ sets all but the $s$ entries of largest magnitude in the input to zero.
For IHT to achieve rapid convergence without degrading as the ambient dimension increases, one needs an estimate of $\bar{L}$, the restricted smoothness constant of \eqref{eq:l0}, which is typically inaccessible. Most works dealing with IHT and variants~\cite{zhou2018,li2016,yuan2018,shen2017} assume this quantity known or perform line search. To our knowledge, the exception to this is  \cite{spolyak} where Polyak step-size is altered to perform well in the high-dimensional setting.
In \cite{spolyak} the authors establish \emph{rate invariance:} the algorithm's convergence rate is independent of the ambient dimension $d$ as long as the achievable statistical precision of the problem stays constant. In the sparse linear regression problem, this implies the same number of iterations are required to achieve a precision of $\mathcal{O}(\frac{s \log (d)}{n})$ regardless of the specific value of $d.$ As shown numerically in \cite{spolyak} leaving Polyak's step-size unaltered in the high-dimensional setting yields step-sizes that are too conservative and that become smaller as the $d$ increases. Many alternatives to Polyak's step-size exist in the literature for low-dimensional problems,  \cite{malitsky2020adaptive, malitsky2024adaptive, latafat2024adaptive}, based on estimating the Lipschitz smoothness constant. However, these will suffer from the same drawback as the standard Polyak step-size. 

For most variants of IHT~\cite{li2016,yuan2018,shen2017}, including \cite{spolyak}, attempting to recover a solution with $s^*$ nonzero entries requires accepting a solution with support size $s = \bar{\kappa}^2 s^*$, where $\bar{\kappa}$ denotes the adequate notion of condition number of \eqref{eq:l0} in the high-dimensional regime of \eqref{eq:l0}. This larger support set leads to a degradation in the final statistical precision and rate. The work of \cite{liu} demonstrates that, by employing alternative thresholding operators, this gap can be narrowed to $s = \bar{\kappa} s^*$. However, their approach  relies on access to $\bar{L}$ or performing line search.


In this work, we extend the adaptive step-size in \cite{spolyak} to allow for other thresholding operators while retaining rate invariance. We show that by suitably choosing thresholding operator, under the adaptive step-size rule \cite{spolyak}, the required support size can be reduced from $s = \bar{\kappa}^2 s^*$ to $s = \bar{\kappa} s^*$. We establish that this not only leads to a sparser solution, but also improves the final achieved statistical precision.

\section{Setup and Background}
We introduce the assumptions on function regularity and the type of thresholding operators we consider additional to HT. Then, we discuss statistical learning models for which our assumptions hold with high probability.
\vspace{-6pt}
\subsection{Function regularity}
The following assumptions generalize the classical notions of strong convexity and $L$-smoothness.
\vspace{-2pt}
\begin{assumption}[RSC \cite{fast}]\label{asp:rscvx}
The objective function $f$ is $(\mu,\tau)$--restricted strongly convex in $\mathbb{R}^d$ if 
\vspace{-2mm}
    \begin{align}\label{eq:rsc}
        &f(\theta_1) - f(\theta_2) - \langle \nabla f(\theta_2),\, \theta_1 - \theta_2 \rangle \notag\\
        &\quad\ge \frac{\mu}{2}\|\theta_1 - \theta_2\|^2 - \frac{\tau}{2}\|\theta_1 - \theta_2\|_1^2, \quad \forall \, \theta_1,\, \theta_2 \in \mathbb{R}^d. \\[-8mm] \nonumber
    \end{align}
\end{assumption}
\vspace{-8pt}
\begin{assumption}[RSS \cite{fast}]\label{asp:rsmooth}
    The objective function $f$ is $(L,\tau)--$restricted smooth in $\mathbb{R}^d,$ if
    \vspace{-2pt}
    \begin{align*}
        &f(\theta_1) - f(\theta_2) - \langle \nabla f(\theta_2),\, \theta_1 - \theta_2 \rangle \\
        &\quad\le \frac{L}{2}\|\theta_1 - \theta_2\|^2 + \frac{\tau}{2}\|\theta_1 - \theta_2\|_1^2,
        \quad \forall\, \theta_1, \theta_2 \in \mathbb{R}^d.
    \end{align*}
\end{assumption}
\vspace{-5pt}
In the high-dimensional setting, objective functions are  not strongly convex, and the Lipschitz smoothness constant scales as $\mathcal{O}(d/n),$ with overwhelming probability \cite{wainwright2019high}. Nevertheless, as demonstrated in \cite{fast}, the restricted variants of these conditions may still hold, with parameters remaining constant or only mildly dependent on $d$ for a wide range of learning problems. We provide examples of fundamental ML problems which exhibit the described behavior Section~\ref{sec:stat}. Observe that if $\theta_{1}, \theta_{2}$ are $s$-sparse and $\mu > 2\tau s$, then Assumption~\ref{asp:rscvx} reduces to strong convexity in the direction $\theta_1 -\theta_2$. The analogous holds for the upper bound yielding a Lipschitz constant scaling with $L,$ $s$ and $\tau$ but  not $d.$ 

 For convenience, we define $\bar{L} = L + 3 \tau s,$ and $\bar{\mu} = \mu - 3 \tau s.$ Further, $\bar{\kappa}$ is formally defined as $\bar{\kappa} = \frac{\bar{L}}{\bar{\mu}}.$

\vspace{-7pt}
\subsection{Sparsifying operator}

A natural choice to obtain sparse solutions based on the formulation \eqref{eq:l0} is the Hard thresholding operator $\Phi^{\mathrm{HT}}_s$ 
\vspace{-3mm}
\begin{align*}
[\Phi^{\mathrm{HT}}_s(\theta)]_i = 
\begin{cases}
\theta_i, & \text{if } i \in S, \\
0,   & \text{otherwise},
\end{cases}\\[-8mm]
\end{align*}
where $S$ contains the indexes of the $s-$largest values of $\theta$ in absolute value.
Under Assumptions~\ref{asp:rscvx} and \ref{asp:rsmooth} IHT requires $s =\mathcal{O}( \bar{\kappa}^2 s^{*}),$ to guarantee linear convergence even if the optimal step-size is chosen. \cite{axiotis} demonstrates that this requirement is not an artifact of the proof but is necessary for IHT. 

In \cite{liu}, the authors establish convergence guarantees of projected gradient descent when swapping the HT operator by others. The authors introduce the notion of \emph{relative concavity}:
\vspace{-5mm}
\begin{align}
\eta_{s^*}(\Phi_s) 
&= \sup \Bigg\{ 
    \frac{\langle y - \Phi_s(z),\, z - \Phi_s(z) \rangle}
         {\| y - \Phi_s(z) \|^2} : \notag\\[-3mm]
&\quad y, z \in \mathbb{R}^d,\ \|y\|_0 \leq s^*,\
     y \neq \Phi_s(z)  \label{eq:concave}
\Bigg\}, \\[-7mm] \nonumber
\end{align}
where $s^* \leq s.$  Note that for any projection $P_C$ onto a convex set $C$ and any $y \in C$,  $\langle y - P_{C}(z), z - P_{C}(z) \rangle \leq 0$ holds $\forall z$. This property no longer holds when $C$ is nonconvex, as is the case for the operators considered in this work. 
In \cite{liu}, it is shown that $\eta_{s^{*}} \left( \Phi^{\mathrm{HT}}_{s} \right) = \frac{\sqrt{s^{*}/s}}{2}$, whereas other thresholding operators have relative concavity $\mathcal{O}(\frac{s^{*}}{s}) $. However, the algorithms discussed in \cite{liu} either require knowledge of the RSM constant to obtain (near) optimal statistical guarantees or a line search mechanism for convergence.
\vspace{-3mm}
\subsection{Statistical models}\label{sec:stat}
\textbf{GLMs:}  
Let \( \theta^{*} \) denote  the ground truth of the statistical model, with $\|\theta^*\|_0 \leq s^*.$
We assume \( x_i \) and \( y_i \) are statistically related via
\vspace{-3mm}
\begin{equation}
\mathbb{P}(y_i \mid x_i, \theta^*, \sigma) = \exp \left\{ \frac{y_i x_i^{T} \theta^* - \psi (x_i^{T} \theta^*) }{c(\sigma)} \right\}, \label{eq:model}
\end{equation}
where \( \sigma > 0 \) is a scale parameter, and \( \psi \) is the cumulant function. $\psi$ is assumed to be infinitely differentiable with $\psi''(t) > 0,$ and uniformly bounded $\forall \,t$. Given this data generation model, we define the objective function
\vspace{-3mm}
\begin{align*}
f(\theta) = \frac{1}{n} \sum_{i=1}^{n} \left( \psi(x_i^{T} \theta) - y_i x_i^{T} \theta \right). \\[-8mm]
\end{align*}   This setting captures a variety of problems, including linear, logistic, and multinomial regression \cite{loh}.   We assume   the feature vectors \( x_i \) are i.i.d. and drawn from a multivariate normal distribution \( N(0, \Sigma) \), where \( \Sigma \) is non-singular. In this setting, when $n < d$, the objective function can not be strongly convex and the Lipschitz smoothness constant scales as $\mathcal{O}(d/n)$ \cite{wainwright2019high}. 
By \cite{loh}, Assumption 2 and variants of Assumption 1 hold with high probability as demonstrated below.

\noindent \textbf{Sparse Linear Regression:} Setting $\psi(t) = t^{2}/2$, \eqref{eq:model} yields sparse linear regression with Gaussian noise.  The  objective function becomes $f(\theta) = \frac{1}{2n}\|y - X\theta\|^{2},$ where $X \in \mathbb{R}^{n \times d}$ aggregates the feature vectors $x_i$ as rows. From \cite{fast}[Lemma 6],  Assumptions~\ref{asp:rscvx} and~\ref{asp:rsmooth} hold  with probability at least \( 1 - e^{-c_{0}n} \) with   \( L = 2 \sigma_{\max}(\Sigma) \), \( \mu = \frac{1}{2} \sigma_{\min}(\Sigma) \), and \( \tau = c_{1} \zeta(\Sigma) \frac{\log d}{n} \), where \( \zeta(\Sigma) = \max_{i=1, \dots, d} \Sigma_{ii} \). Here \( c_{0} \) and \( c_{1} \) are universal constants. 
 Assumption~\ref{asp:rscvx} does not generally hold for other GLMs. We introduce a weaker variant for GLMs such as logistic  and multinomial regression \cite{loh}. 
\begin{assumption}[weak RSC]\label{asp:weak}
A function $f$ is $(\mu,\tau)$--weakly-restricted strongly convex in $\mathbb{R}^d$ if for any $\theta_1, \theta_2 \in \mathbb{R}^d$, 
\vspace{-3mm}
    \begin{equation}\label{eq:glm_rsc}
    \begin{aligned}
    &f(\theta_1) - f(\theta_2) - \langle \nabla f(\theta_2), \theta_1-\theta_2 \rangle \geq \\
    &\begin{cases}
    \frac{\mu}{2}\|\theta_1-\theta_2\|_2^2 - \frac{\tau}{2}\|\theta_1-\theta_2\|_1^2, & \|\theta_1-\theta_2\|_2 \leq 1, \\
    \|\theta_1-\theta_2\|_2\!\left(\tfrac{\mu}{2} - \tfrac{\tau}{2}\tfrac{\|\theta_1-\theta_2\|_1^2}{\|\theta_1-\theta_2\|_2^2}\right), & \|\theta_1 -\theta_2\|_2 > 1.
    \end{cases}
    \end{aligned}
    \end{equation}
\end{assumption}
\vspace{-3mm}
Note that \eqref{eq:glm_rsc} is always implied by \eqref{eq:rsc}.

\noindent \textbf{Sparse Logistic Regression:} Setting $\psi(t) = \log (1 + e^{t}) $, \eqref{eq:model} reduces to logistic regression. For $y_{i} \in \left\{ 0, 1 \right\} $, the distribution is $p(y_{i}) = p_{i}^{y_{i}} (1 - p_{i})^{1-y_{i}}$, with $p_{i} = \frac{e^{x_{i}^{\top} \theta^{*}}}{1 + e^{x_{i}^{\top} \theta^{*}}} $. By \cite[Appendix D.1]{loh} and \cite[Proposition 2]{nega}, it can be shown that the objective function satisfies the RSS and weak RSC conditions  with probability at least $1 - c_{1} d^{-1} - c_{2} e^{-n} $, where $\mu $ and $L $ depend only on $\Sigma $, and $\tau = c_{3} \frac{\log d}{n} $, with $c_{1}, c_{2}, c_{3} >0 $ being universal constants.
\vspace{-3mm}
\section{Main Results}
\vspace{-3mm}
Our work \cite{spolyak} adapts Polyak's step size to perform well in the high-dimensional regime. We establish that even as the ambient dimension grows, the required number of iterations required to achieve (near) optimal statistical precision remains the same. However, we require $s = \mathcal{O}(s^{*}\bar{\kappa}^2),$ leading to losses both in terms of sparsity and statistical performance. To alleviate this, we generalize Sparse Polyak  to allow for alternative sparsifying operators $\Phi_s$ while still employing an adaptive step-size rule. We provide the proposed algorithm in Algorithm~\ref{algo:iht} followed by our main result.
\vspace{-1mm}
\begin{algorithm}
\caption{Sparse Polyak}
\label{algo:iht}
\begin{algorithmic}[1]
\State \textbf{Input:} Function $f$, target function value $\widehat{f},$ sparsity parameter \( s \),  number of iterations $T$
\State \textbf{Initialize:} \( \theta_0 \in \mathbb{R}^d, \) with  \(\|\theta_0\|_0 \leq s\)
\For{t = 0 to \( T -1\)}
    \State Compute step-size \(\gamma_t = \frac{\max\{f(\theta_t) - \widehat{f},0\}}{5 \| \hs(\nabla f(\theta_{t})) \|^2}\)
    \State Update: \( \theta_{t+1} = \Phi_{s}\left( \theta_t - \gamma_t \nabla f(\theta_t) \right) \)
\EndFor
\State \textbf{Output:} \( \theta_T \)
\end{algorithmic}
\end{algorithm}
\vspace{-3mm}

\begin{theorem}\label{thm:1}
    Let \(\{\theta_t\}_{t \geq 1}\) denote the iterates generated by Algorithm~\ref{algo:iht}. Suppose  \(f\) is convex and satisfies Assumptions~\ref{asp:rscvx} and~\ref{asp:rsmooth}. Let \(\widehat{\theta}\) be any \(s^{*}\)-sparse vector such that \(f(\widehat{\theta}) = \widehat{f}\). Assume $s/s^{*}$ is sufficiently large to guarantee that \(\bar{\mu} > 0\), and $\eta_{s^*}(\Phi_s) \leq 1/4$.
    Then, for any \(\theta_t\) such that \(\|\theta_t - \widehat{\theta}\|^2 \geq \frac{36\|\hs(\nabla f(\widehat{\theta}))\|^2}{\bar{\mu}^2}\)  we can guarantee
    \[
    \|\theta_{t+1} - \widehat{\theta}\|^2 \leq \left(1 - \frac{1}{40 \bar{\kappa}}+ 4\eta_{s^*}(\Phi_s)\right) \|\theta_t - \widehat{\theta}\|^2.
    \]
    If Assumption~\ref{asp:weak} holds instead of Assumption~\ref{asp:rscvx}, set $\gamma_t = \frac{\max\{f(\theta_t) - \widehat{f},0\}}{5 \| \hts(\nabla f(\theta_{t})) \|^2}$. If, further $\bar{\mu} \geq 4 \| \hts(\nabla f(\widehat{\theta})) \| 
$ and $\theta_{0} = 0,$ we can guarantee that $ \| \theta_{t+1} - \widehat{\theta} \|^{2} \leq$
    \[
    \begin{cases}
    \left( 1 - \frac{1}{80\bar{\kappa}} + 4\eta_{s^*}(\Phi_s) \right) \| \theta_t - \widehat{\theta} \|^{2} & \| \theta_t - \widehat{\theta} \| \leq 1,\\
             \left( 1 - \frac{1}{80 (4\|\widehat{\theta}\|^2+1))\bar{\kappa}^{2}} + 4\eta_{s^*}(\Phi_s) \right) \| \theta_t - \widehat{\theta} \|^{2}        & \| \theta_t - \widehat{\theta} \| > 1.
    \end{cases}
    \]

 In both cases, if for some $T \geq 0$ there holds \(\|\theta_T - \widehat{\theta}\|^2 < \frac{36\|\hs(\nabla f(\widehat{\theta}))\|^2}{\bar{\mu}^2}\), then
    \begin{align*}
    \|\theta_t - \widehat{\theta}\|^2 \leq \left(1 + 4\eta_{s^*}(\Phi_s)\right) \frac{36\|\hs(\nabla f(\widehat{\theta}))\|^2}{\bar{\mu}^2}
    \quad \forall t \geq T
    \end{align*}
\end{theorem}
Theorem~\ref{thm:1} guarantees a contractive relation among iterates when $-1/(40\bar{\kappa}) + 4\eta_{s^{*}}(\Phi_{s}) < 0$, and the optimal rate is attained when $\eta_{s^{*}}(\Phi_{s}) \leq c \bar{\kappa}^{-1}$ for some $c > 0$. From \eqref{eq:concave}, $\eta_{s^{*}}$ decreases with increasing $s$. Thus $s$ can be chosen sufficiently large to guarantee contractiveness. Then, the iterates reach a ball centered at $\widehat{\theta}$ with radius $\frac{36\|\hs(\nabla f(\widehat{\theta}))\|^2}{\bar{\mu}^2}$, which grows with $s$. Therefore, it is desirable to select the thresholding operator that yields the smallest $s$ while ensuring fast convergence. We provide the proof for Theorem~\ref{thm:1} in \cite{arxiv}.

For the HT operator $\eta_{s^{*}}(\Phi_{s}^{\mathrm{HT}})=\frac{\sqrt{s^{*} / s}}{2} $, which implies that the optimal rate is attained when $s = \mathcal{O}(s^* \bar{\kappa}^2),$ recovering the results in \cite{spolyak}.
Theorem~\ref{thm:1} extends the main result of \cite{spolyak} from hard thresholding to any thresholding operator with bounded relative concavity. Among them, the reciprocal thresholding (RT) operator can weaken the requirement on $s$ and guarantee better accuracy.
The RT \cite{liu} is defined as
\[
\Phi^{\mathrm{RT}}_{s}(\theta)_{i} =
\begin{cases}
\text{sign}(\theta_{i}) \left( \frac{1}{2} \left| \theta_{i} \right| + \frac{1}{2} \sqrt{\left| \theta_{i} \right|^{2} - \tau^{2}} \right), & i\in S,\\
                 0  & i\notin S,
\end{cases}
\]
where $\tau = \max_{i\notin S} \left| \theta_{i} \right| $ is the magnitude of the $s+1$ largest entry of $x .$ Observe that the RT operator guarantees hard sparsity while at the same time, shrinking the coordinates within the support.
RT has a significantly smaller relative concavity,  bounded by $\frac{s^{*} /s}{\min \left\{ 1, 4(1-(s^* /s)) \right\}}$, c.f. \cite[Lemma 6]{liu}. Incorporating this into Theorem~\ref{thm:1} immediately yields the following corollary.

\begin{corollary}\label{coro:operators}
Under the same setting as Theorem~\ref{thm:1}, we compare the convergence behavior of different thresholding operators. In particular, for each operator we characterize (i) the lower bound on the iterate sparsity level $s$ that ensures the contractive relation
\[
\|\theta_{t+1} - \widehat{\theta}\|^2 \leq \left(1 - \frac{1}{80 \bar{\kappa}}\right) \|\theta_t - \widehat{\theta}\|^2,
\]
and (ii) the corresponding best achievable accuracy in terms of $\|\theta_t - \widehat{\theta}\|$. The results are summarized in Table~\ref{tab:operators}.
\end{corollary}
\vspace{-15pt}
\begin{table}[h]
\centering
\caption{Comparison of sparsity requirements and achievable accuracy for different thresholding operators.}
\label{tab:operators}
\begin{tabular}{lcc}
\toprule
\textbf{Operator} & \textbf{$s$} & \textbf{Best accuracy of $\|\theta_t - \widehat{\theta}\|$} \\
\midrule
HT & $\mathcal{O}(s^{*}\,\bar{\kappa}^2)$ & $\mathcal{O}\!\left(\tfrac{\bar{\kappa}^{2}\|\hxs(\nabla f(\widehat{\theta}))\|^{2}}{\bar{\mu}^{2}}\right)$ \\
RT & $\mathcal{O}(s^{*}\,\bar{\kappa})$ & $\mathcal{O}\!\left(\tfrac{\bar{\kappa}\|\hxs(\nabla f(\widehat{\theta}))\|^{2}}{\bar{\mu}^{2}}\right)$ \\
\bottomrule
\end{tabular}
\end{table}
The use of the RT operator improves the previous requirement for IHT from $s \geq \mathcal{O}(s^{*} \bar{\kappa}^{2}) $ to the milder condition $s \geq \mathcal{O}(s^{*}\bar{\kappa}) $. The achievable precision improves by a factor $\bar{\kappa}$. In the following, we provide statistical guarantees for linear regression and general GLMs.

\begin{corollary}[Linear Regression]\label{coro:lg}
Consider the sparse linear regression problem described in Section~\ref{sec:stat}. Further suppose that the sample size is sufficiently large so that \(\bar{\mu} > 0\). Further, assume the design matrix \(X \in \mathbb{R}^{n \times d}\) is normalized such that \(\| X_j / \sqrt{n} \| \leq C\) for all \(j = 1, \dots, d\). Let \(\{\theta_t\}_{t \geq 0}\) be the sequence of iterates produced by Algorithm~\ref{algo:iht} when applied to the sparse linear regression problem. Assume the sparsity parameter satisfies \(s \geq 320\, \bar{\kappa} s^*\), and $\widehat{f} = f(\theta^*)$. Then, with probability at least \(1 - e^{-c_0 n} - \frac{2}{d}\), there holds:

\textbf{(i)}If \(\|\theta_t - \theta^*\|^2 \geq \frac{288 C^2 \sigma^2 s \log d}{n \bar{\mu}^2}\), the iterates exhibit contraction toward \(\theta^*\), i.e., $\|\theta_{t+1} - \theta^*\|^2 \leq \left(1 - \frac{1}{80 \bar{\kappa}}\right) \|\theta_t - \theta^*\|^2.$

\textbf{(ii)} Let \(T\) denote the first iteration at which \(\|\theta_{T} - \theta^*\|^2 < \frac{288 C^2 \sigma^2 s \log d}{n \bar{\mu}^2} \). Then for all \(t \geq T\), the iterates remain confined in a neighborhood of \(\theta^*\): $\|\theta_t - \theta^*\|^2 \leq \left(1 + \frac{1}{80\bar{\kappa}}\right) \frac{288 C^2 \sigma^2 s \log d}{n \bar{\mu}^2}.$
\end{corollary} 
\vspace{-2mm}
\begin{corollary}[Generalized Linear Models]\label{coro:glm}
Let \( \{ \theta_t \}_{t \geq 0} \) denote the iterates generated by Algorithm~\ref{algo:iht} when applied to the generalized linear models described in \eqref{eq:model}. Assume that \( \psi \) is infinitely differentiable with \( \psi''(t) > 0 \) and uniformly bounded for all \( t \in \mathbb{R} \). Set the step size rule according to
\vspace{-2mm}
\begin{align*}
\gamma_t = \frac{\max\{f(\theta_t) - \widehat{f},0\}}{5 \| \hts(g_t) \|^2}. \\[-4mm] 
\end{align*}
Define \( R:= \| \theta^* \|^2 \), and \( R_0:= 4R+1 \). Assume we set $\widehat{f} = f^*,$ $\theta_0 = 0,$ and suppose \(s \geq 640 R_0 \bar{\kappa}^2 s^*\). Further, assume the sample size is large enough to ensure $\bar{\mu} > 5 c_{0} \sqrt{\frac{ 2 s \log d}{n}} $ for some universal constant $c_0 > 0$.

Then, with probability at least \( 1 -  \frac{c_{1}}{d} - c_{2} e^{-n} \), where $c_1, c_2 >0$ are universal constants, we guarantee that at most three modes of convergence:
(i) If $\|\theta_t - \theta^{*}\| \geq 1 $, then
\[
\|\theta_{t+1} - \theta^*\|^2 \leq \left(1 - \frac{1}{160 R_0 \bar{\kappa}^2} \right)\|\theta_t - \theta^*\|^2.
\]
(ii) If $  1 >\|\theta_t - \theta^{*}\|^2 \geq \frac{6c_0^2s \log d}{n\bar{\mu}^2}$, then
\[
\|\theta_{t+1} - \theta^*\|^2 \leq \left(1 - \frac{1}{160 \bar{\kappa}} \right)\|\theta_t - \theta^*\|^2.
\]
(iii) If for some $T>0$ we have, $\|\theta_T - \theta^{*}\|^2 \leq \frac{36c_0^2s \log d}{n\bar{\mu}^2}$, then we guarantee that for any $t \geq T$, $\|\theta_T - \theta^{*}\|^2 \leq \left(1 + \frac{1}{160 R_0 \bar{\kappa}^2} \right)\frac{36c_0^2s \log d}{n\bar{\mu}^2}$.
\end{corollary}
In both cases, we achieve (near) optimal statistical precision at a rate that is independent of the ambient dimension. This is achieved without requiring knowledge of $\bar{L}.$ While we do require an estimate of $f(\theta^{*})$ we can establish similar results with only a lower bound \cite{spolyak}. Alternatively, $f(\theta^{\star})$ can be estimated as $f(\theta^{\star}) = \mathcal{O}(\sqrt{\frac{s^{*}\log(d)}{n}})$ as done in the low-dimensional case in \cite{ren2022towards}.

\section{Numerical Experiment}

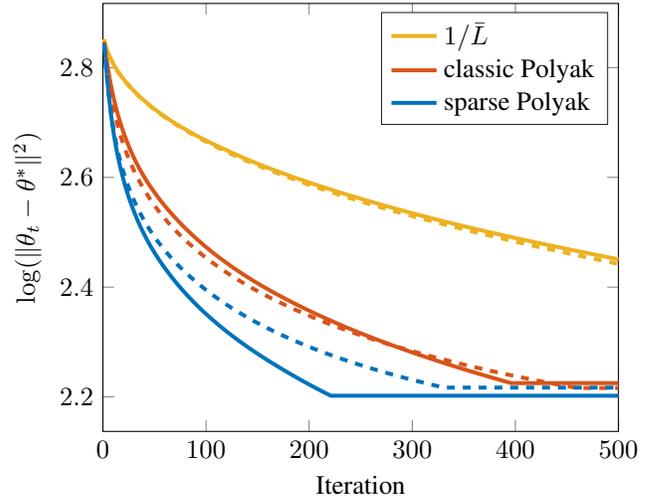
\begin{figure}
    \centering
    \definecolor{mycolor3}{rgb}{0.00000,0.44700,0.74100}%
\definecolor{mycolor2}{rgb}{0.85000,0.32500,0.09800}%
\definecolor{mycolor1}{rgb}{0.92900,0.69400,0.12500}%

\begin{tikzpicture}
\begin{axis}[
    xlabel={Iteration},
    ylabel={$\log(\|\theta_t - \theta^*\|^2)$},
    axis background/.style={fill=white},
    legend style={legend cell align=left, align=left, draw=white!15!black},
    xmin=0,
    xmax=500,
    title style={font=\bfseries},
    title={Logistic Regression},
]

\addplot[color=mycolor1,line width=1.5pt] table {data1.txt};
\addplot[color=mycolor2,line width=1.5pt] table {data2.txt};
\addplot[color=mycolor3,line width=1.5pt] table {data3.txt};

\addplot[color=mycolor1,line width=1.5pt, dashed] table {data4.txt};
\addplot[color=mycolor2,line width=1.5pt, dashed] table {data5.txt};
\addplot[color=mycolor3,line width=1.5pt, dashed] table {data6.txt};

\legend{{$1/\bar{L}$}, {classic Polyak},{sparse Polyak},{},{},{}}
\end{axis}
\end{tikzpicture}
    \label{fig:htrt}
    \caption{We use dashed line for HT operator with $s=500$, and solid line for RT operator with $s=300$.}
\end{figure}

In this section we perform sparse logistic regression. We set $d =5000$, $s^* = 300,$ and $n = \lceil 5 s^* \log d \rceil.$ $X \in \mathbb{R}^{d \times n}$ is generated to reflect a time-series structure with a correlation parameter $\omega = 0.5$. For each column index $j \in \{1, \dots, n\}$, we generate a sequence of i.i.d. standard normal variables $\varepsilon_1, \dots, \varepsilon_{d}$, and set $x_{1,j} = \varepsilon_1 / \sqrt{1 - \omega^2}$. Then, $x_{t,j} = \omega x_{t,j} + \varepsilon_t$ for $t \in \{2, \dots, d\}.$ The true parameter $\theta^*$ is created by sampling each entry from $\mathcal{N}(0, 1)$, and selecting at random which $s^*$ values remain. $y_i$ is generated following the model \eqref{eq:model} with the choice described in Section~\ref{sec:stat} for logistic regression. The sparsity parameter $s$ for the HT and RT operators is selected by grid search over the set $\{300, 400, 500, 600, 700\}$. Figure~\ref{fig:htrt} illustrates the performance of projected gradient descent with a fixed step size rule, classic and sparse Polyak with both HT and RT. In all cases, the parameter $s$ yielding best final accuracy is chosen. For the fixed-step rule we set $\gamma_t = 1/\hat{L}$ with 
$
\hat{L} = \lambda_{\max}(\Sigma)\,\left(\frac{3}{4}+\frac{2s + s^{*}}{10s} \right),
$
following the formulation in \cite[Appendix D.1]{loh}.


Sparse Polyak consistently achieves faster convergence. Further, incorporating the RT operator leads to iterates with fewer nonzero entries and thus a more accurate final solution. Further, because $\bar{\kappa}$ itself degrades with increasing $s,$ we see an increase in speed when using the RT.
\vspace{-3mm}
\section{Conclusion}
\vspace{-2mm}
In this paper we extended the results in \cite{spolyak}  to achieve both rate invariance and improved statistical precision. This was done by allowing for a larger variety of thresholding operators with improved relative concavity. We provided both theoretical and numerical results.


\clearpage
\section{Appendix}
\subsection{Proof of Theorem~\ref{thm:1}}
\begin{proof}
For simplicity, we let $g_{t} = \nabla f(\theta_{t}) $, and $\widehat{g} = \nabla f(\widehat{\theta}) $.
Let $\mathcal{S}_{t}$ be the support of $\theta_t$, 
$\widehat{\mathcal{S}}$ that of $\widehat{\theta}$, and 
$\widehat{\mathcal{S}}_{t} = \mathcal{S}_{t} \cup \widehat{\mathcal{S}}$. 
Define $\widetilde{\theta}_{t+1} = \theta_t - \gamma_t g_t$.

We begin with the decomposition
\begin{align*}
\|\theta_{t+1} - \widehat{\theta}\|^2
&= \big\| \theta_{t+1} - [\widetilde{\theta}_{t+1}]_{\widehat{\mathcal{S}}_{t+1}}
+ [\widetilde{\theta}_{t+1}]_{\widehat{\mathcal{S}}_{t+1}} - \widehat{\theta} \big\|^2 \\
&= \| \theta_{t+1} - \left[ \widetilde{\theta}_{t+1} \right]_{\widehat{S}_{t+1}} \|^{2} \\
 &\quad + 2 \langle \widehat{\theta} - \left[ \widetilde{\theta}_{t+1} \right]_{\widehat{S}_{t+1}}, \left[ \widetilde{\theta}_{t+1} \right]_{\widehat{S}_{t+1}} - \theta_{t+1} \rangle\\
 &\quad + \| \left[ \widetilde{\theta}_{t+1} \right]_{\widehat{S}_{t+1}} - \widehat{\theta} \|^{2}\\
&= \|\theta_{t+1} - [\widetilde{\theta}_{t+1}]_{\widehat{\mathcal{S}}_{t+1}}\|^2
  \\
 &\quad + 2\langle \widehat{\theta} - \theta_{t+1}, [\widetilde{\theta}_{t+1}]_{\widehat{\mathcal{S}}_{t+1}} - \theta_{t+1} \rangle \\
&\quad + \|[\widetilde{\theta}_{t+1}]_{\widehat{\mathcal{S}}_{t+1}} - \widehat{\theta}\|^2 - \|[\widetilde{\theta}_{t+1}]_{\widehat{\mathcal{S}}_{t+1}} - \theta_{t+1}\|^2 \\
&= 2\langle \widehat{\theta} - \theta_{t+1}, \widetilde{\theta}_{t+1} - \theta_{t+1} \rangle
 + \|[\widetilde{\theta}_{t+1}]_{\widehat{\mathcal{S}}_{t+1}} - \widehat{\theta}\|^2 \\
&\quad - \|[\widetilde{\theta}_{t+1}]_{\widehat{\mathcal{S}}_{t+1}} - \theta_{t+1}\|^2.
\end{align*}

By the definition of $\eta_{s^*}(\Phi_s)$, this yields
\[
\|\theta_{t+1} - \widehat{\theta}\|^2
\le 2\eta_{s^*}(\Phi_s) \|\theta_{t+1} - \widehat{\theta}\|^2
 + \|[\widetilde{\theta}_{t+1}]_{\widehat{\mathcal{S}}_{t+1}} - \widehat{\theta}\|^2.
\]
Given that $\eta_{s^*}(\Phi_s) \leq 1/4$, it further implies 
\begin{equation}\label{eq:base_thm1}
\|\theta_{t+1} - \widehat{\theta}\|^2
\le \big(1 + 4\eta_{s^*}(\Phi_s)\big)
\|[\widetilde{\theta}_{t+1}]_{\widehat{\mathcal{S}}_{t+1}} - \widehat{\theta}\|^2.
\end{equation}
Equation \eqref{eq:base_thm1} indicates that the deviation introduced by hard thresholding is controlled: the iterate \(\theta_{t+1}\) remains close to \(\widetilde{\theta}_{t+1}\) in terms of their distances to \(\widehat{\theta}\).

From the proof of \cite[Theorem 1]{spolyak}, when \(\|\theta_t - \widehat{\theta}\|^2 \geq \frac{36\|\hs(\widehat{g})\|^2}{\bar{\mu}^2}\), the sparse Polyak step size is bounded below by $\mathcal{O}(\frac{1}{\bar{L}})$. Specifically, under Assumption~\ref{asp:rscvx}, choosing the sparse Polyak step size ensures
\[
\|[\widetilde{\theta}_{t+1}]_{\widehat{\mathcal{S}}_{t+1}} - \widehat{\theta}\|^2
\le \left(1 - \frac{1}{40\bar{\kappa}}  \right)\|\theta_t - \widehat{\theta}\|^2.
\]
Combining \eqref{eq:base_thm1}, we have
\[
\|\theta_{t+1} - \widehat{\theta}\|^2
\le \left(1 - \frac{1}{40\bar{\kappa}} + 4\eta_{s^*}(\Phi_s) \right)\|\theta_t - \widehat{\theta}\|^2.
\]
From the proof of \cite[Theorem 1]{spolyak}, when \(\|\theta_t - \widehat{\theta}\|^2 < \frac{36\|\hs(\widehat{g})\|^2}{\bar{\mu}^2}\), we still have $ \|[\widetilde{\theta}_{t+1}]_{\widehat{\mathcal{S}}_{t+1}} - \widehat{\theta}\|^2
\le \|\theta_t - \widehat{\theta}\|^2,$ which guaranties that $\|\theta_{t+1} - \widehat{\theta}\|^2
\le \big(1 + 4\eta_{s^*}(\Phi_s)\big)
\|\theta_t - \widehat{\theta}\|^2 $ always holds. Combining these two cases, we complete the proof when Assumption~\ref{asp:rscvx} holds. 

The only change when replacing Assumption~\ref{asp:rscvx} with Assumption~\ref{asp:weak} arises in the case $\| \theta_t - \widehat{\theta} \| > 1 $. Adopting the same argument as in \cite[Lemma 4]{spolyak}, we obtain
\[
\|[\widetilde{\theta}_{t+1}]_{\widehat{\mathcal{S}}_{t+1}} - \widehat{\theta}\|^2
\leq \| \theta_t - \widehat{\theta} \|^{2} - 2 \gamma_t (f(\theta_t) - f(\theta^{*})) + 5 \gamma_t^{2} \| \hts(g_t) \|^{2}.
\]
With the chosen step size, this simplifies to
\[
\|[\widetilde{\theta}_{t+1}]_{\widehat{\mathcal{S}}_{t+1}} - \widehat{\theta}\|^2
\leq \| \theta_t - \widehat{\theta} \|^{2} - \frac{(f(\theta_t) - f(\widehat{\theta}))^{2}}{5 \| \hts (g_t) \|^{2}}.
\]
Moreover, under the condition  $\bar{\mu} \geq 4 \| \hts(\widehat{g}) \|$, Assumption~\ref{asp:weak} yields

\[
    f(\theta_{t}) - f(\widehat{\theta}) \geq \frac{\bar{\mu}}{2} \| \theta_{t} - \widehat{\theta} \| - \| \hts (\widehat{g}) \| \| \theta_{t} - \widehat{\theta} \| \geq \frac{\bar{\mu}}{4} \| \theta_{t} - \widehat{\theta} \| .
\]
This indicates that \( f(\theta_t) - f(\widehat{\theta}) \) grows linearly with \( \|\theta_t - \widehat{\theta}\| \).
 In contrast, under strong RSC, the function gap typically grows quadratically with the distance, i.e., \( f(\theta_t) - f(\widehat{\theta}) \propto \|\theta_t - \widehat{\theta}\|^2 \).
Substituting this into equation~\eqref{eq:base_thm1}, we obtain
\begin{align*}
\| \theta_{t+1} - \widehat{\theta} \|^{2} &\leq \left( 1 + 4\eta_{s^*}(\Phi_s) - \frac{\bar{\mu}^{2}}{80 \| \hts (g_{t}) \|^{2}} \right) \| \theta_{t} - \widehat{\theta} \|.
\end{align*}
Finally, by applying the same reasoning as in the proof of \cite[Corollary 5]{spolyak}, the result follows. 

\end{proof}
The proof of Theorem~\ref{thm:1} proceeds in two steps:
(i) establishing the potential expansion effect introduced by the hard thresholding operator, captured in equation~\eqref{eq:base_thm1};
(ii) deriving the contractive relation between $\widetilde{\theta}_{t+1} - \widehat{\theta} $ and $\theta_{t} - \widehat{\theta} $, which follows naturally from the gradient descent update. Compared with \cite[Theorem 1]{spolyak}, the only distinction lies in the argument for part (i).

\subsection{Proof of Corollary~\ref{coro:lg} and~\ref{coro:glm}}
These results follow directly from Theorem~\ref{thm:1}, while the theoretical guarantees for the RSS and RSC conditions are provided in \cite[Corollaries 4 and 5]{spolyak}.

\end{document}